\documentclass[letterpaper, 10 pt, conference]{ieeeconf}  
\pdfoutput=1
\IEEEoverridecommandlockouts                              
\usepackage{gensymb}
\usepackage{cclicenses, graphicx}
\usepackage{amsmath}
\usepackage{array}

\usepackage[keeplastbox]{flushend}

\newcolumntype{C}[1]{>{\centering\arraybackslash}m{#1}}

\title{\LARGE \bf EMG Pattern Classification to Control a Hand Orthosis for\\Functional Grasp Assistance after Stroke}

\author{Cassie Meeker$^{1}$, Sangwoo Park$^{1}$, Lauri Bishop$^{2}$, Joel Stein$^{2,3}$ and Matei Ciocarlie$^{1,3}$%
\thanks{*This work was supported in part by the Columbia-Coulter Translational
Research Partnership, NSF grant IIS-1526960 (part of the National Robotics Initiative) and ONR Young Investigator Program award N00014-16-1-2026.}%
\thanks{$^{1}$Department of Mechanical Engineering, Columbia University, New York, NY 10027, USA.}%
\thanks{\hspace{-3mm}{\tt\small \{cgm2144, sp3287, matei.ciocarlie\}@columbia.edu}}%
\thanks{$^{2}$Department of Rehabilitation and Regenerative Medicine, Columbia University, New York, NY 10027, USA. {\tt\small \{lb2413, js1165\}@cumc.columbia.edu}}%
\thanks{$^{3}$Co-Principal Investigators}
}

\begin{document}

\maketitle
\thispagestyle{empty}
\pagestyle{empty}

\begin{abstract}
Wearable orthoses can function both as assistive devices, which allow the user to live independently, and as rehabilitation devices, which allow the user to regain use of an impaired limb.  To be fully wearable, such devices must have intuitive controls, and to improve quality of life, the device should enable the user to perform Activities of Daily Living.  In this context, we explore the feasibility of using electromyography (EMG) signals to control a wearable exotendon device to enable pick and place tasks.  We use an easy to don, commodity forearm EMG band with 8 sensors to create an EMG pattern classification control for an exotendon device.  With this control, we are able to detect a user's intent to open, and can thus enable extension and pick and place tasks.  In experiments with stroke survivors, we explore the accuracy of this control in both non-functional and functional tasks. Our results support the feasibility of developing wearable devices with intuitive controls which provide a functional context for rehabilitation.

\end{abstract}

\section{Introduction}

Wearable devices are an attractive alternative to other robotic
rehabilitation therapies that, traditionally, require therapist
supervision provided in a clinical setting, and take place in a
non-functional context. Therapy is more likely to be effective when
training is distributed in smaller but more frequent
aliquots~\cite{shea1991} and when training includes performing actual
Activities of Daily Living (ADLs)~\cite{krakauer2006}. Wearable
devices can provide both of these advantages.

For devices to be wearable in a functional context, they need
intuitive, user-driven controls. We are developing an
electromyography (EMG) controlled exoskeleton hand orthosis for stroke
patients. This is a step toward user-controlled take-home
orthotic devices to help perform functional tasks.

\begin{figure}[t]
\centering
\begin{tabular}{c}
\includegraphics[width=1.0\linewidth]{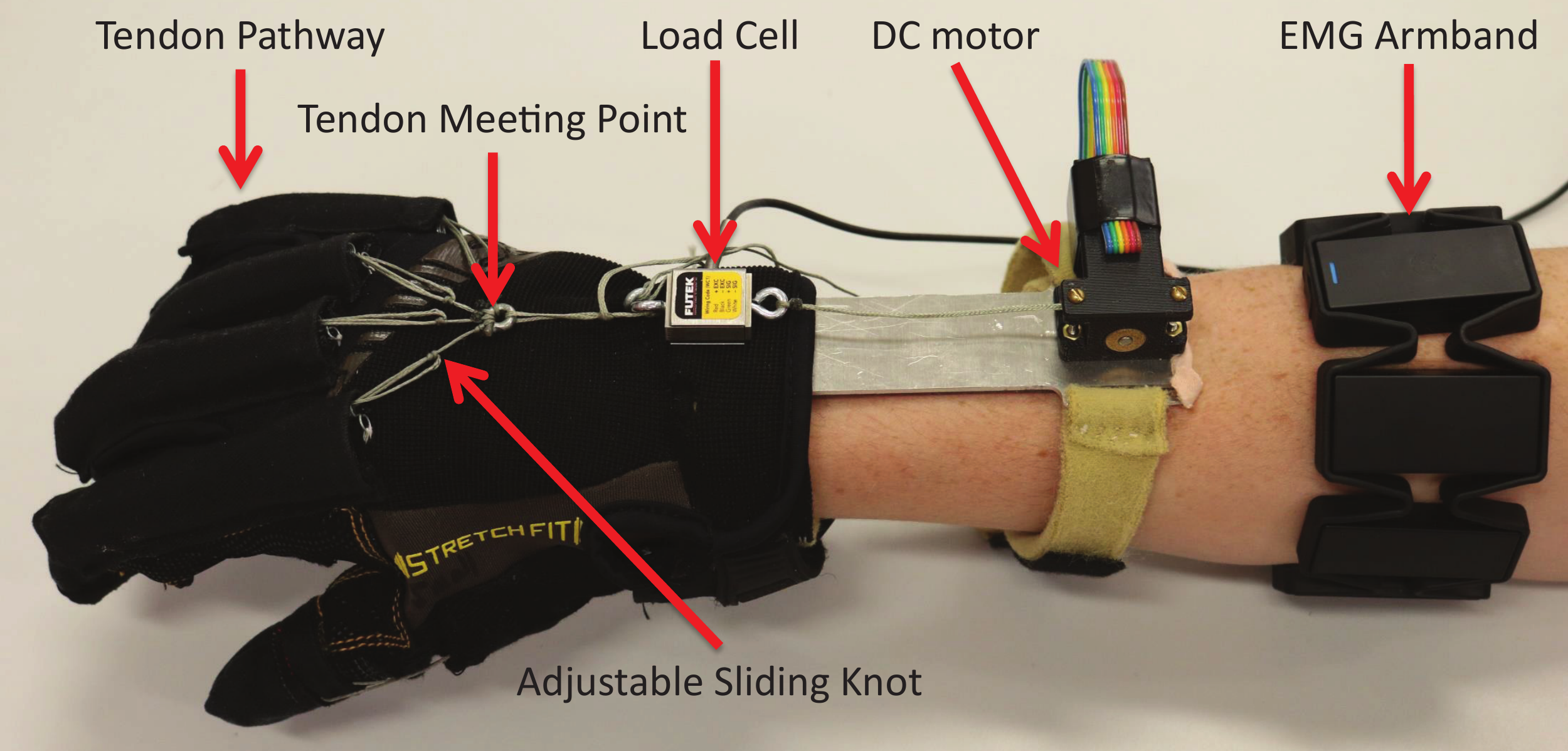}
\end{tabular}
\caption{Prototype hand orthosis in extension configuration with EMG armband.}
\label{fig:orthosis}
\vspace{-15pt}
\end{figure}

In previous work, we presented an exotendon orthosis - a soft glove
with guided tendons driven by linear electric actuators to elicit
desired movement patterns~\cite{park2016}. Here, we create and test a
control mechanism for the exotendon device based on surface
EMG. Fig.~\ref{fig:orthosis} shows this exotendon device, including
the EMG armband which provides control input.

Our approach uses pattern classification - identifying patterns in EMG
signals from the entire forearm - to determine user intention. In
this study, we apply EMG pattern classification to determine the
user's intention either to open or to close the hand, and use this
signal to produce physical movement assisted by the orthosis. While we
do not measure cognitive load, the short training sessions that enable
the user to operate the device suggest that using signals from the
same muscles that drive an unimpaired hand is an intuitive control
mechanism for an orthosis. Overall, we present a complete mechanism
comprising hardware and algorithms to:
\begin{itemize}
\item Detect a motor impaired user's intention to execute specific
  hand movements based on forearm EMG.
\item Use an encapsulated commodity EMG sensing suite without needing precise
  sensor positioning. This is in contrast to medical grade EMG sensors
  placed on a specific muscle by a trained professional, which do not
  allow in-home use outside of direct medical supervision.
\item Physically elicit the desired movement pattern in stroke
  patients. We do not study EMG in isolation: we
  combine the control with a real orthosis and show it enables
  functional grasping in our target population. The presence of the
  physical device alters the EMG data obtained during operation; our
  approach is designed to cope with this phenomenon.
\end{itemize}
\noindent To the best of our knowledge, no existing hand orthotic system
has concurrently demonstrated all of these
characteristics. The key to our approach is pattern classification,
which enables the use of commodity EMG armbands. Commodity armbands 
have the potential to allow complete portability as
well as user-directed, in-home use. \textit{Pattern classification of
  functional movements for stroke subjects has so far not been studied
  in conjunction with a physical orthosis that enables
  grasping.} Our experience indicates it is important to study
these components together, as we aim to progress towards fully
user-driven execution of complete tasks.

\section{Related Work}

Proposed control methods for orthoses include brain control interfaces
~\cite{buch2008}, bilateral control~\cite{kawasaki2007}, therapist/user 
driven orthoses~\cite{takahashi2008} and EMG control. In this study, we 
use an EMG-based control mechanism. This control method 
has been used successfully for orthoses of the knee~\cite{hassani2013}, upper 
arm~\cite{vaca2013}, and wrist~\cite{khokhar2010}, as well as the hand. 

A number of proposed EMG controls for hand orthoses include the use of 
bilateral muscles~\cite{lucas2004} or bicep muscles~\cite{lucas2004,
 dicicco2004}. We have chosen to focus on ipsilateral muscles of the
forearm, aiming for more intuitive control for the user, and also
leaving the bilateral hand free to operate independently.
Furthermore, most forearm muscles are used specifically to manipulate
the hand, whereas bicep muscles are not.

Hand orthoses using ipsilateral forearm EMG controls are largely
developed with either the goal of hand function improvement or grasp
assistance~\cite{maciejasz2014}. A number of studies have shown the
ability to process EMG data in order to predict hand position
~\cite{mulas2005, kavya2015, chen2009}; the armband we use to
collect EMG signals has a similar built-in ability to extract hand control
signals based on typical EMG patterns of a healthy
user. None of the above studies, nor the armband, demonstrate the feasibility
of using their control scheme in conjunction with an orthosis that enables the 
execution of functional tasks. Our experience indicates that 
methods which predict hand motions in absence of an orthosis need significant 
adaptation to be effective when an orthosis is present (Section
~\ref{sec:exotraining}). Our study presents a method for training the EMG control in the presence of an orthosis to better predict hand position while the orthosis 
is operating.

In EMG-driven controls that are developed with orthoses to enable
functional tasks~\cite{tong2010, ho2011, ochoa2011}, 2 sensors are
placed by trained experimenters on specific pairs of muscles. The
control then uses a threshold to determine when to open and close the
orthosis. Our control does not require placement on specific muscles
to function, resulting in a much easier donning process. Our commodity 
sensors are fully encapsulated and wearable, with no
separate amplification or power electronics. Both of these features
are important for building wearable orthotics that patients will one
day be able to use at home without direct medical
supervision. Furthemore, instead of looking only at pairs of muscles
and thresholding activation signals (as in the above studies), we
investigate patterns within the EMG signal of the entire forearm. This
allows for future development of the EMG control to include multiple
and varied hand positions. It has been shown that pattern
classification can identify multiple hand positions in
stroke patients~\cite{lee2011}; therefore, we predict our control
approach will be able to grow and develop with our orthosis design as,
in future studies, we continue to enable more hand positions in stroke
patients.

As stated above, pattern classification of functional movements for stroke 
subjects has been studied with the ultimate goal of
controlling an orthosis~\cite{lee2011}. However, to our knowledge, this kind 
of classification has not been studied in conjunction with an 
orthosis that enables functional movements. Our experience indicates 
that EMG controls for stroke patients need to be developed in conjunction
with orthoses. We explore the feasibility of using pattern classification 
while simultaneously using an orthosis to enable hand extension.

EMG control in the field of prosthetics is well documented and can
provide insights for EMG control of orthotics. Proportional EMG
control~\cite{fougner2012} and pattern recognition based EMG
control~\cite{powell2013} of prosthetics are paradigms also used in
orthotics. Prosthetic EMG control of individual finger
motions~\cite{kanitz2011, bitzer2006} and prosthetic EMG control which
determines force and grasp type~\cite{castellini2009} have been
explored. Control optimization studies have informed both signal
processing~\cite{ajiboye2005,englehart2003} and control
hierarchy~\cite{cipriani2008} of prostheses.

In prosthetics, there is also the question of how to train the device
to perform optimally. The 2 main approaches are system training and
user training. System training is adapting the control to be more
accurate and can involve gains, thresholds, computer-guided training
or bilateral training to provide ground truth for the
system~\cite{fougner2012}. User training is teaching the user to
produce control signals that are easily distinguishable for the system
and, in the context of EMG controls, involves teaching the user how to
create consistent and distinguishable muscle
patterns~\cite{powell2013}. We believe that this latter approach can
also prove valuable for the types of signals we use here, and plan to
apply it in future iterations.

\section{Exotendon Device}

Our EMG-based control approach is implemented and tested on a complete
hand orthosis device, which we briefly describe here. In prior work,
we presented a four fingered orthosis with two separate 1-degree of freedom 
(DOF) tendon configurations\cite{park2016}. Here, we experiment with the 
configuration assisting with extension, often a difficult task for
stroke patients because of the commonly observed impairment pattern of 
spasticity, which is excessive involuntary flexion. We chose to 
explore the prototype which enables extension because extension is 
essential for functional grasping. Extension is achieved by
applying extension torques on the fingers through an exotendon network 
pulled by a DC motor.

Mechanical components are split into two modules: a forearm piece and a glove with a 
tendon network. The modules are connected via eyerings on both sides 
of the load cell (Futek, FSH00097) to facilitate donning 
(Fig.~\ref{fig:orthosis}). With a therapist, donning takes approximately 5 minutes. 
The device weighs 135 grams. 

The forearm piece is composed of an aluminum splint and a DC
motor. The splint constrains wrist movement to efficiently transmit
external torques to the fingers. The splint is angled either at 30
degrees, considered functional wrist pose~\cite{natasha2003}, or at 0
degrees for the patients who cannot extend the wrist due to
spasticity. A DC motor (Pololu corporation, 210:1 Low-Power Micro
Metal Gearmotors) with a 15 mm/s maximum travel speed and a 80N peak
force is mounted on the splint. Motor specifications are chosen to
prevent dangerous tendon force levels without taking up space to
improve wearability. A Proportional-Integral-Derivative (PID)
position control is implemented to drive the motor. The motor's range
of motion is determined at the beginning of clinical tests, depending
on hand size.

The glove with a tendon network has tendons guided from the heads of
the middle phalanges through raised pathways to a meeting point on the
back of the hand. The tendons on each finger are attached to a cloth
ring on the middle phalanges rather than on the fingertips to avoid
finger hyperextension. The tendons on all digits, except for the
thumb, are routed on the dorsal side of the glove. The thumb needs a
special routing scheme since it exhibits different movement patterns
from the other digits; the thumb tendon is routed from the proximal
phalanx head to the metacarpal joint, then wraps clockwise around the
wrist to the eyering on the load cell. The tendons on all digits are
tied with sliding adjustable knots to allow better fit for different
finger lengths.

The DC motor which pulls the exotendon network is driven by the
EMG-based control that is the main focus of this study. Once the EMG
control determines the user's intention to open or close the hand, it
sends a command to the DC motor mounted on the splint. The motor then
extends or retracts the tendon network to allow the user to open or
close the hand.

\section{EMG Control} \label{control}

EMG patterns of the hemiparetic forearm are often altered after a
stroke event~\cite{dewald1995}. This study is based on the assumption
that these altered EMG patterns can still be used to control a hand
orthosis; as control using forearm EMG sensors has a number of
compelling characteristics. EMG-based control requires the same type
of muscle activation as pre-stroke extension, which should make the
control intuitive and place a low cognitive load on the user.
Additionally, using ipsilateral EMG control leaves the other hand free
to participate in the grasping task or to perform a different task.

Beyond altered signals however, EMG control of an orthosis for a
stroke patient is difficult because of additional phenomena, such
as spasticity and abnormal coactivation relationships between
muscles~\cite{dewald1995}. As such, many orthoses that enable pick
and place collect signals from only two muscles~\cite{tong2010,
 ho2011, ochoa2011}, with each muscle controlling a direction of the
orthosis, often using a threshold based on the subject's maximum
voluntary contraction. In these approaches, the subject must fully
extend or close before the orthosis will move in the other
direction. We aim to develop an exotendon device that responds
immediately to a signal change from the control, throughout the range
of motion of the user's hand. The user's ability to end extension
allows more natural grasping for smaller objects as well as the option
to change grasping tasks mid-motion.

One of the key tenets of our approach is to rely on signals from a
multitude of sensors placed around the
forearm. Unlike simple intensity thresholding, which is effective
for a single sensor precisely located on a specific muscle, 
pattern classification identifies patterns in the complete set of
signals from the sensors. This approach has three main benefits:

\begin{enumerate} 
\item It enables the use of commodity sensors. Even though the
 quality of the EMG signal from commodity sensors is lower than
 medical grade sensors, we compensate for signal quality with sensor
 quantity. Pattern classification provides an image of the overall
 EMG signal in the entire forearm instead of trying to isolate a high
 quality signal from specific muscles.
\item It eliminates the need to search for specific muscles with exact
 sensor placement. Pattern recognition examines EMG signals from the
 entire forearm. Studies have suggested that when electrodes are
 placed around the entire forearm, targeted and untargeted placement
 of EMG electrodes result in similar classification
 accuracies~\cite{farrell2008}. Throughout our experiments, the only
 effort to position our EMG sensors was placing one of the sensors
 on the dorsal side of the arm. Even with this untargeted approach,
 we were still able to use pattern classification with good
 accuracy. The flexibility in sensor placement means that donning our
 control unit does not require a therapist, or even a basic
 understanding of forearm anatomy. For a device that is designed for
 take-home use in mind, this is an extremely desirable quality.
\item It allows for the possibility of an orthosis with more DOFs. 
 Current orthoses look at two specific muscles, a
 flexor and an extensor. The flexor controls the close motion of the
 orthosis and the the extensor controls the open motion. Pattern
 classification allows for the recognition of more complex muscle
 motions, which could control different DOFs of the
 orthosis~\cite{powell2013}.
\end{enumerate}

To acquire the EMG signal, we use the Myo Armband from Thalmic Labs.
It has 8 EMG sensors and 8 IMUs, which can indicate the orientation
and acceleration of the device. In this study, we only
use the EMG sensors; however, the IMU sensors could be useful for future 
control iterations.

\subsection{Pattern Classification}

Our pattern classification algorithm seeks to take the 8-dimensional
raw EMG data from the 8 Myo sensors and identify patterns that
correspond to certain desired hand motions. Our current
algorithm only identifies hand opening and closing, but we
hope to incorporate more complex patterns into the
classification scheme in future iterations.

We collect raw EMG data from the Myo Armband at a rate of 50Hz. 
At time $t$, we collect the EMG signals $e_j^t$ from the sensors and 
assemble them into a data vector $\psi_t$:
\begin{equation}
\psi_t = (e_1^t \dots e_8^t)
\end{equation}

We define the desired hand state at time $t$ as $H_t \in \{O,C\}$,
where $H_t=O$ corresponds to the intent to open the hand and $H_t=C$
is the intent to close the hand. While training, ground
truth data $H_t^g$ is provided by the experimenter who gives the
subject verbal commands to open or close the hand. The training period 
is around 45 seconds - allowing the experimenter to command the
user to try to open and close the hand twice. Although this
training time is short, we receive a large quantity of data
points ($\sim$2,400) which we use to establish 
patterns in the EMG with our classifier.

Raw EMG signals are used as the features for the classifier. 
Although many pattern classifiers require extraction of time-domain
features, we receive our data at 50 Hz, so this
would be impractical. Our results in Section~\ref{experiments} show 
our classifier is robust enough that it does not need to extract time-domain
features to classify intention with high accuracy. 

Our first order goal is to predict $H_t$ based on $\psi_t$ (we will
further process this result as explained in the next sections). We use
a random forest classifier trained on the ground truth data described
above to make this prediction. A random forest classifier is an
ensemble machine learning method created from a combination of tree
predictors~\cite{breiman2001}. Because of the random nature of the
bootstrap sampling used to create our classifier, the number of decision 
trees in the forest classifier and the decision trees themselves change 
with every training iteration. Despite the underlying randomness, our 
classifiers for all subjects still achieve high accuracy. 

We denote the random forest classifier function as:
\begin{equation}
CLAS(\psi_t) = p_t^O \in [0,1]
\end{equation}
where $p_t^O$ is the probability that $H_t=O$ (at time $t$, the user's
intention is to open the hand). The converse probability that the
user's intent is to close the hand is simply $p_t^C = 1 -
p_t^O$. We filter and use this result as described in the next
section.

\subsection{Output Processing}
We collect raw EMG data $\psi_t$ at a rate of 50Hz. However, the time
scale for hand opening and closing and for pick and place tasks
is much lower frequency than the rate at which data is collected, so
classifying individual data points correctly is not as crucial as
correctly identifying a hand motion. To identify these motions, we
assume hand posture does not change with high frequency, which
allows us to filter and process the probabilities returned by the
classifier.

While filtering raw EMG signals is a common technique, we chose
instead to apply our filter to the results of the classifier. We
compute filtered probabilities at time $T$ as:
\begin{eqnarray}
\hat{p}_T^O &=& \texttt{MEDIAN}(p_t^O), t \in [T-0.5s, T] \\
\hat{p}_T^C &=& \texttt{MEDIAN}(p_t^C), t \in [T-0.5s, T]
\end{eqnarray}
The 0.5s median filter increases transition delays, but helps
eliminate spikes and spurious predictions. 0.5s was chosen because 
shorter filters resulted in spurious classification errors. Despite
the delay, our subjects reported no noticeable delay between intention 
initiation and device movement. We note that, as a result of filtering, 
generally $ \hat{p}_T^O + \hat{p}_T^C \neq 1$.

To produce the final output for our control, we compare
$\hat{p}_T^O$ and $\hat{p}_T^C$ against two threshold
levels, $L^O$ and $L^C$ respectively. If $\hat{p}_T^O \geq L^O$, then
the controller issues an ``open'' command (retract the
tendon). If $\hat{p}_T^C \geq L^C$, then the controller
issues a ``close'' command (extend the tendon). If neither condition
is met, no new command is issued and the orthosis continues
executing the command from the previous step. The values of $L^O$ and
$L^C$ are set manually by the experimenter for each subject after
completing training data collection, then kept constant throughout all
tests. The thresholds are set with subject
feedback such that the control is responsive, but there are no
spurious errors during sustained hand commands.

\subsection{Training with the Exotendon Device}
\label{sec:exotraining}

\begin{table}
\vspace{2mm}
\centering
\begin{tabular}{c|ccc}
Device State & \multicolumn{3}{c}{Subject Instruction}\\
&Open&Relax&Close\\\hline
\\[-2.5mm]\hline
\\[-3mm]
Tendon extended 	& O & C & C \\
Tendon retracting 	& O & & C \\
Tendon retracted 	& O & C & C 
\end{tabular}
\caption{Training protocol and assigned labels. For each combination
 of instruction given to the subject and state of the exotendon
 device, the table shows the ground truth label $H_t^g$ assigned to
 EMG data. Training begins with the tendon extended and the subject
 asked to relax (top row, middle column) and proceeds in
 counter-clockwise fashion.}
\vspace{-25pt}
\label{tab:train}
\end{table}

The most straightforward method for generating training data to use
with the classifier described above would be to simply instruct the
user to attempt to open or close the hand, and label the resulting
data accordingly. However, we quickly found that this simple procedure
is flawed for multiple reasons. First, for stroke patients, we found
that the default ``relaxed'' hand state (attempting to neither open nor
close) still produces a strong, subject-specific EMG signal. The classifier 
would displayed a tendency to label this
signal as either open or close, unless we provided explicit training
data illustrating the difference. Second, we also found that physical
interaction with the orthosis itself altered the EMG patterns: for the
same user intention, signals recorded with the tendon fully retracted
(assisting in hand opening) differed from those recorded with the
tendon extended.

We address both of these issues through our training protocol and
collection of labeled training data. Specifically, we design our training protocol as follows:
\begin{itemize}
\item We instruct the subject to attempt three
 hand poses: open, closed, and relaxed. For data collected
 during both closed and relaxed intents, we assign a ground truth
 label $H_t^g=C$, corresponding to a closed hand. Since our 
 target population comprises patients with spasticity, this more
 closely mimics the subjects' natural state. This means
 that, for the orthosis to provide assistance, we must be
 detecting an active attempt by the user to open their
 hand. Being conservative in when to send a command to retract the
 tendon (and thus actively open the hand) reduces the risk of holding
 the hand open for longer than desired and causing
 discomfort. We note that one disadvantage is that continuous effort
 from the subject can lead to muscle fatigue, especially if the
 subject exerts great strain to provide an open signal.
\item For all three user intents (open, close, relaxed) we collect
 training data in different states of the exotendon device, namely
 with the tendon fully extended, fully retracted, or moving between
 states. The training procedure is as follows. We 
 instruct the subject to relax, with the tendon fully extended. We
 ask the subject to attempt to open the hand, with the tendon still
 fully extended. As the subject continues trying to open, the
 experimenter commands the tendon to retract, opening the hand. Once
 the tendon is fully retracted, we instruct the subject first to
 relax, then to attempt to close the hand. The experimenter then
 commands the tendon to extend, allowing the hand to close. Finally,
 the subject is told to relax. This procedure, and the
 ground truth labels assigned at every phase, are summarized in
 Table~\ref{tab:train}.
\end{itemize}
The result of this training procedure is a labeled ground truth
dataset covering combinations of user intent and device state. We use
this dataset to train the classifier described above; at run
time, the output of the classifier produces a command for
the exotendon device as detailed in the previous section.

\section{Experiments and Results} \label{experiments}

Testing was performed with 4 stroke survivors, 1 female and 3 male. Subjects 
showed right side hemiparesis following a stroke event at least 2.5 years 
prior and had a spasticity level between 1 and 3 on
the Modified Ashworth Scale (MAS). Table~\ref{demographic_table} shows 
clinical scores for all subjects. Testing was approved by the
Columbia University Internal Review Board, and performed in a clinical
setting under the supervision of Physical and/or Occupational
Therapists.

\begin{table}[]
\vspace{2mm}
\centering
\caption{Subject Demographic and Clinical Information}
\label{demographic_table}
\begin{tabular}{C{0.7cm}|C{.4cm}|C{.6cm} C{.6cm} C{.6cm}|C{.6cm} C{.6cm} C{.6cm}}
 	&       	& \multicolumn{3}{c|}{MAS Extensor Score}     	& \multicolumn{3}{c}{MAS Flexor Score}    \\ 
Subject & Age		& Elbow  	& Wrist 	& Finger	& Elbow		& Wrist		& Finger   \\ \hline
A       & 60      	& 2         	& 1      	& 1         	& 2		& 2		& 2           \\ 
B       & 39      	& 1         	& 0      	& 0          	& 2		& 3		& 3            \\ 
C       & 80      	& 0         	& 0      	& 1          	& 1		& 0		& 2            \\ 
D       & 66      	& 2          	& 0        	& 0           	& 2		& 2		& 1            \\ 
\end{tabular}
\vspace{-5mm}
\end{table}

We asked each subject to don the Myo and the exotendon device with the
assistance of the supervising therapist. Training the EMG control as 
described in Section~\ref{control} was performed for every session. The 
trained classifier was then used throughout the entire session. In real 
deployment, we would like the classifier to be robust enough to take the 
armband off and put it back on. We predict this is possible if the armband orientation on the forearm is consistent; however, this was not explored 
here because of the possibility of EMG patterns changing as the patient 
underwent rehabilitation. We aim to do this in future iterations of the control.

After training, each subject performed 4 experiments:
\begin{enumerate}
\item \textbf{EMG control without the device operating:} This
  experiment determined if the EMG signal in the
  hemiparetic forearm arm was strong enough to indicate the subject's
  intention to open or close. Without the device operating, there was
  little hand movement, but we still were able to determine the user's
  intention.
\item \textbf{EMG control with the device operating:} This experiment
  determined whether EMG control, in conjunction with our exotendon
  orthosis, could enable hand extension. With the device on, the Myo
  sends raw EMG signals to the classifier, which predicts intent and
  sends a command based on intent to the motor, which retracts or
  extends the tendon to move the hand and enable extension. Because
  this enables extension, it requires the training protocol described
  in Section~\ref{control}. The subject's forearm was at rest on the table.
\item \textbf{EMG control during pick and place:} This experiment
  determined whether the exotendon device, in conjunction with
  EMG control, could enable pick and place. The exotendon device
  enabled hand extension but the forearm was no longer supported by
  the table.
\item \textbf{Button control during pick and place:} This experiment
  provided a baseline control comparison for the EMG control. A push
  button is attached to the device's motor and can be used to retract
  and extend the tendon. Pushing down and holding the button opens the
  glove until the hand is fully extended. Releasing the button at any
  point of the extend cycle causes the tendon to be released
  immediately and allows the hand to relax. The subject used the
  button with the non-affected hand to activate the device and
  complete pick and place tasks.
\end{enumerate}

Experiments were performed at a pace comfortable for the subject
and breaks were given between experiments.

Our result reports include two metrics: prediction accuracy and
correctly predicted events. Prediction accuracy is defined as the
percentage of individual data points $\psi_t$ predicted by the
classifier to be the same as ground truth. However, we believe that
the more important metric is the ability to correctly execute a
complete, meaningful hand motion, such as opening or closing. We
attempt to capture this using the number of correctly predicted
events. An event is defined as a change in intention
signal, and a correctly predicted event means a predicted event which
occurs within 850 ms of the ground truth event, with no incorrect
classifications until the next event. The 850ms 
allowed for lag introduced by the median filter and allowed 
the subject intitate the action after a verbal command.
Success for this metric was if the EMG control did as well as the baseline
button control.

\subsection{EMG control without the device operating}

To collect the training set, the subject was asked to try to open and
close the hemiparetic hand, with the understanding that the fingers
likely would not extend, but that the EMG signal would change as
different actions were attempted. (Note that this is the only
condition in which we did not use the training protocol described in
Section~\ref{sec:exotraining}.) The testing set was collected in the same
way as the training set. The subject's hand did not move, but the
classifier was able to predict the subject's intention by the EMG
signals.

\begin{figure}[t]
\vspace{2mm}
\centering
\begin{tabular}{r}
\includegraphics[trim=5mm 0mm 5mm 0mm,clip=true,width=1\linewidth]{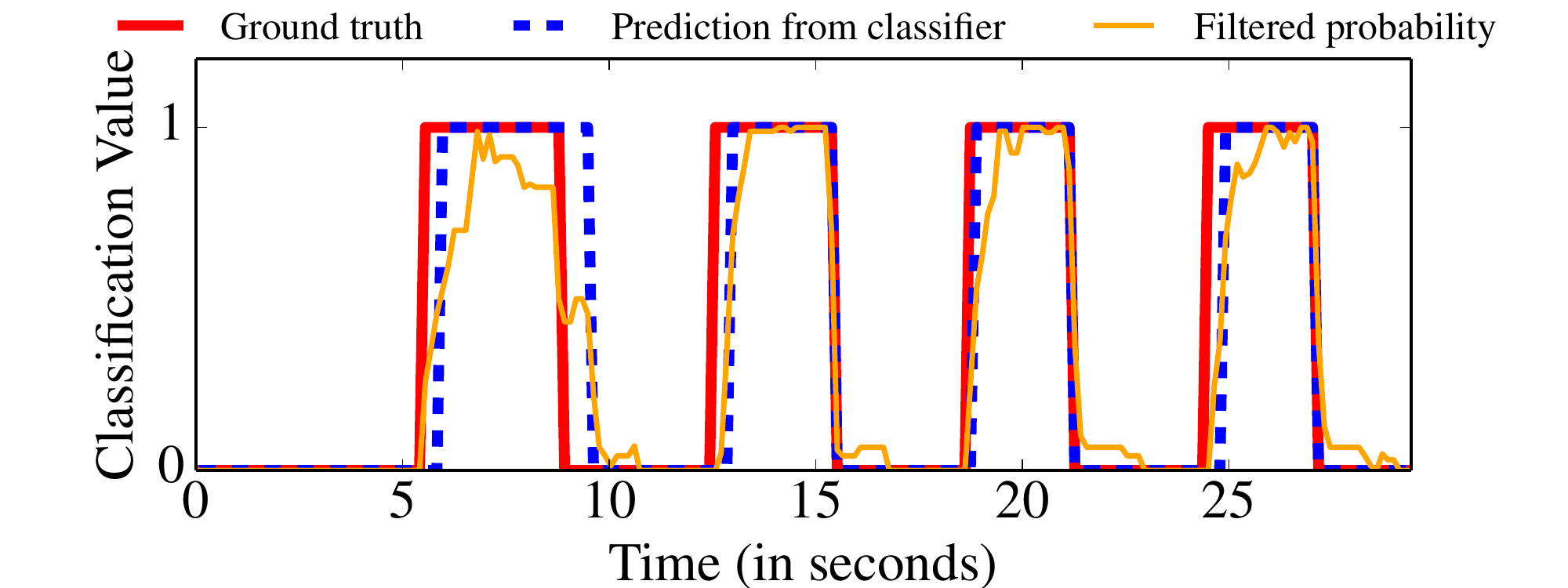}\\
\vspace{-5mm}
\includegraphics[trim=5mm 0mm 5mm 4mm,clip=true,width=1\linewidth]{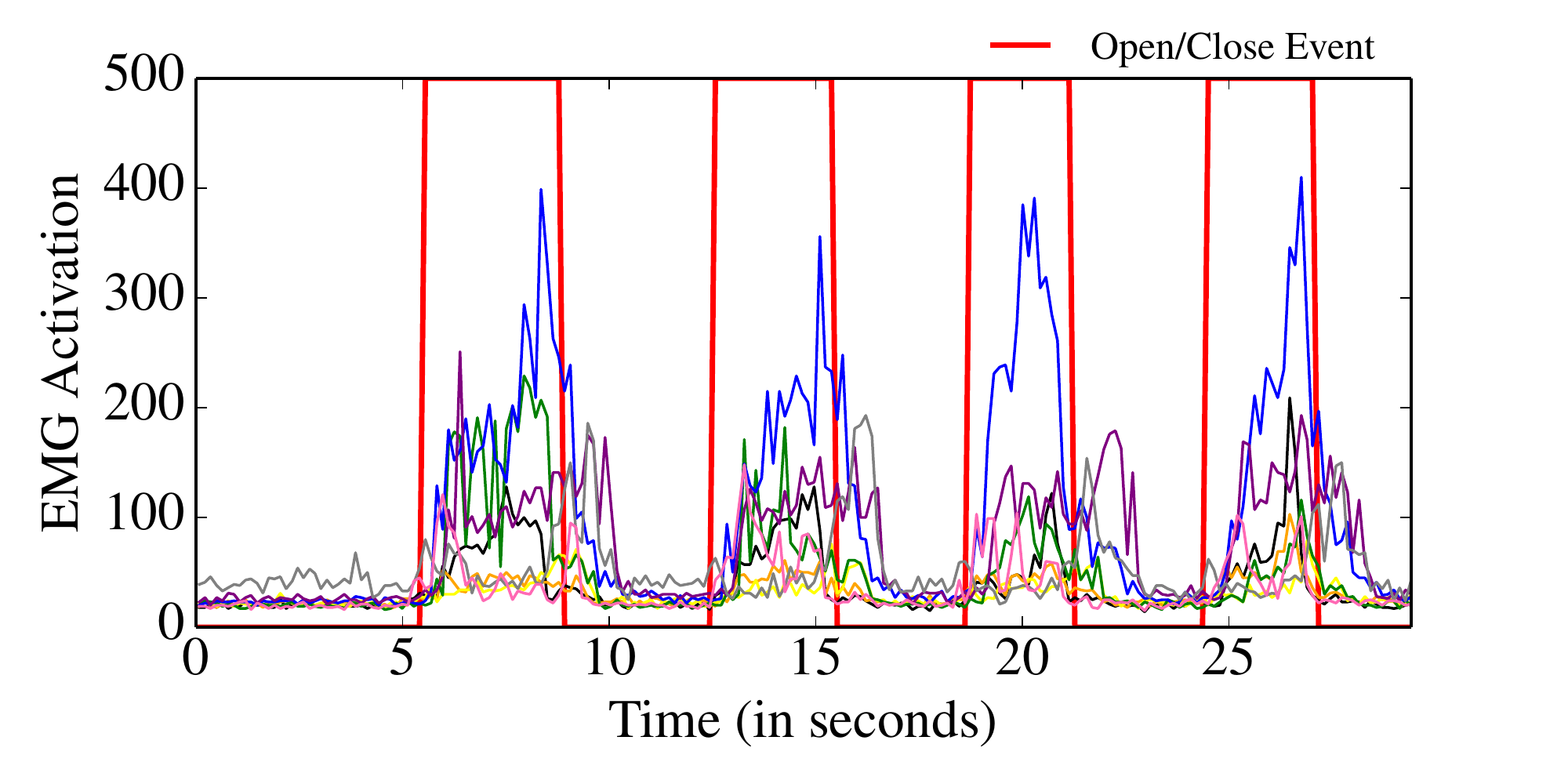}\\
\end{tabular}
\caption{\textbf{Top}: Prediction of classifier (dotted blue), ground truth (solid red) and 
	 filtered probability before thresholding (solid orange) vs. time for
	 Subject A without the device operating. Classification value of 1: open 
	 intention, classification value of 0: no open intention. \textbf{Bottom}:
	 Raw EMG of Subject A and open close events which correspond to the
	 top graph.}
\label{fig:without_device_results}
\vspace{-3mm}
\end{figure}

The classifier for Subject A had an accuracy of 85.2\% and
correctly predicted 11 of 18 events. The intention for Subject B
was predicted with 90.1\% accuracy and 10 of 16 events were
correctly predicted. The classifier for Subject C had an
accuracy of 93.6\% and correctly predicted 12 of 14 events. Subject D's
intention was predicted with a 82.2\% accuracy and 4 
of 10 events were predicted. Fig.~\ref{fig:without_device_results}
shows ground truth, prediction results and non-thresholded filtered probability vs. 
time of Subject A, as well as the raw EMG which is classified. 
See Table~\ref{nonfunctional_table} for a summary of the results.

\begin{table}[]
\centering
\vspace{2mm}
\caption{Results for Non-Functional Motions}
\label{nonfunctional_table}
\begin{tabular}{C{0.9cm}|C{1.4cm}C{1.3cm}|C{1.4cm}C{1.3cm}}

	& \multicolumn{2}{c|}{Without Device}      & \multicolumn{2}{c}{With Device}         \\ 
Subject & \% Accuracy & Correct Events & \% Accuracy & Correct Events \\ \hline
A       & 85.2\%      & 11/18                      & 93.6\%      & 16/18                      \\ \hline
B       & 90.1\%      & 10/16                      & 83.4\%      & 4/16                       \\ \hline
C       & 93.6\%      & 12/14                      & 90.9\%      & 9/11                       \\ \hline
D       & 82.2\%      & 4/10                       & N/A         & N/A                        \\ 
\end{tabular}
\vspace{-4mm}
\end{table}

\begin{figure}[t]
\vspace{2mm}
\centering
\begin{tabular}{r}
\includegraphics[trim=5mm 0mm 5mm 0mm,clip=true,width=1\linewidth]{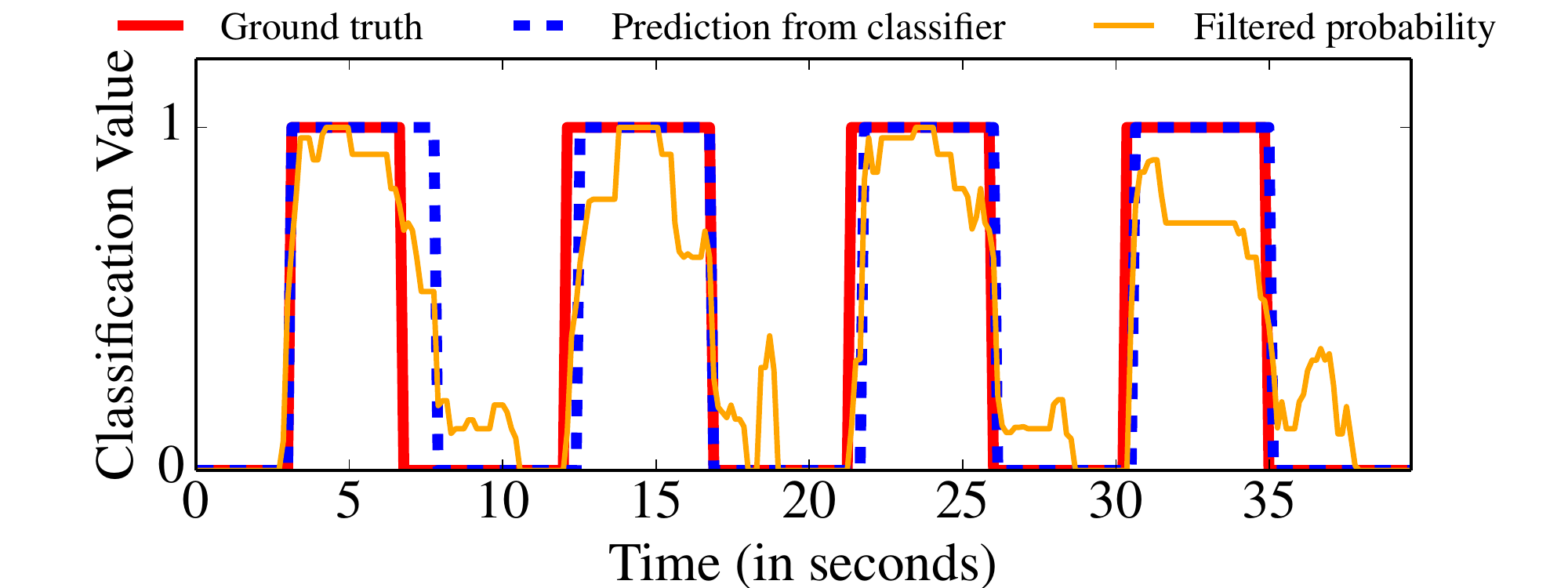} \\
\includegraphics[trim=5mm 0mm 5mm 4mm,clip=true,width=1\linewidth]{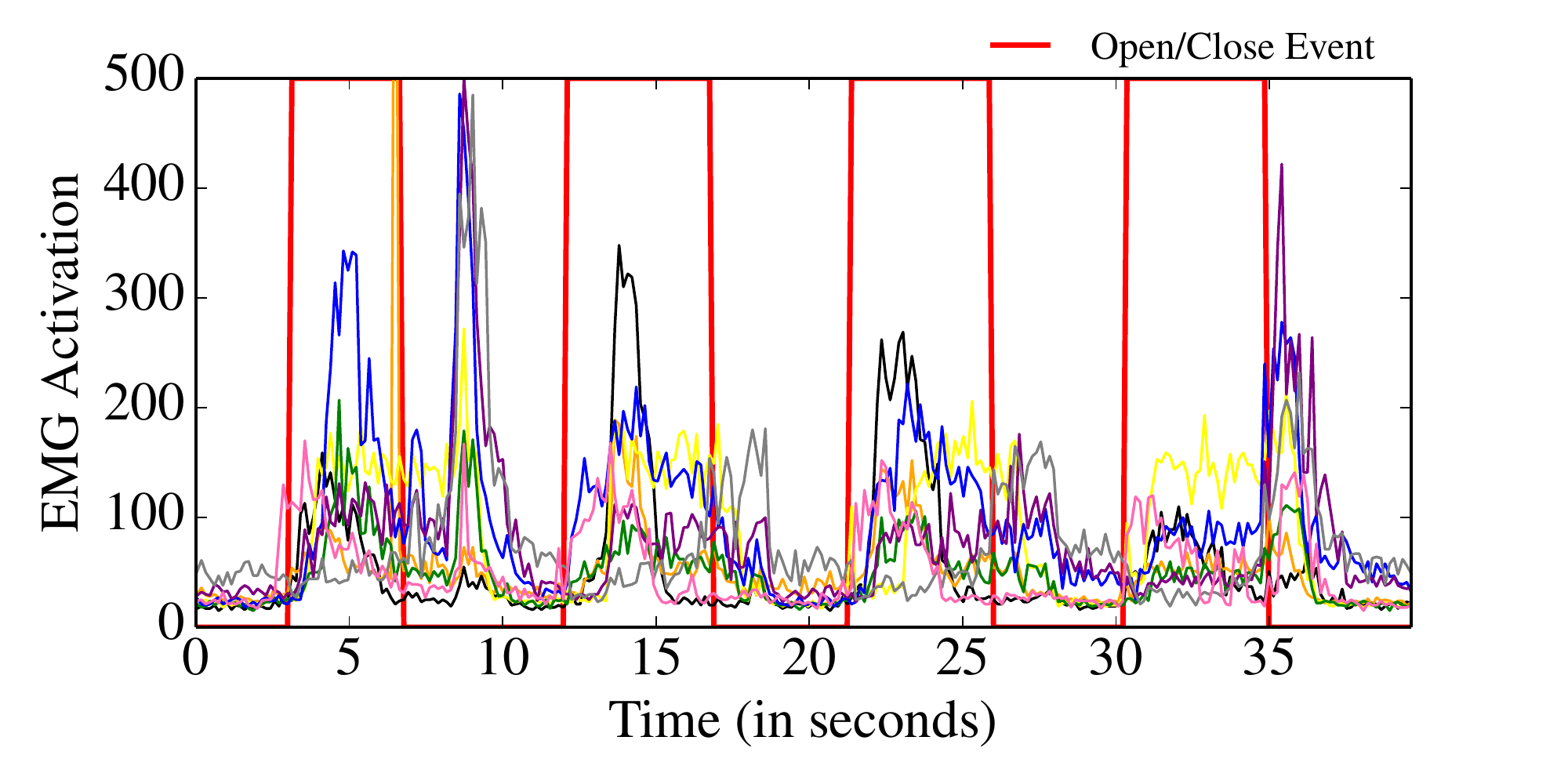}\\
\end{tabular}
\vspace{-7mm}
\caption{\textbf{Top}: Prediction of classifier (dotted blue), ground truth (solid red) and 
	 filtered probability before thresholding (solid orange) vs. time for
	 Subject A with the device operating. Classification value of 1: open 
	 intention, classification value of 0: no open intention. \textbf{Bottom}:
	 Raw EMG of Subject A and open close events which correspond to the
	 top graph.}
\label{fig:with_device_results}
\vspace{-2mm}
\end{figure}

\subsection{EMG control with the device operating}

In this section, the device was functioning to extend the
hand, so training used the protocol from
Section~\ref{sec:exotraining}.

The testing set was collected as the subject was asked to try to open
and relax the hemiparetic hand while resting the hand on the
table. If the classifier detected the subject was attempting to open,
the exotendon device would retract the tendon and the subject's hand
would extend. If the intention to open was absent, the device allowed
the hand to close.

Subject D was not included in these results because of subject
fatigue. The classifier for Subject A had a prediction accuracy of
93.6\% and correctly predicted 16 of 18 events. The classifier for
Subject B had an accuracy of 83.4\% and correctly predicted
4 of 16 events. Subject C had an accuracy of 90.9\% and
the classifier correctly identified 9 of 11 events. The
finger flexor MAS score for Subject B was higher than for Subjects A and C.
This could explain why Subject B's accuracy and correct
event prediction are lower. An plot of the ground truth and
the prediction results vs. time of Subject A, as well as the raw EMG 
which is classified, can be found in Fig.~\ref{fig:with_device_results}. 
See Table~\ref{nonfunctional_table} for a summary of these results.

\begin{figure*}[t]
\vspace{2mm}
\setlength{\tabcolsep}{1mm}
\centering
\begin{tabular}{ccccc}
\includegraphics[trim=15mm 0mm 0mm 15mm,clip=true,width=0.18\linewidth]{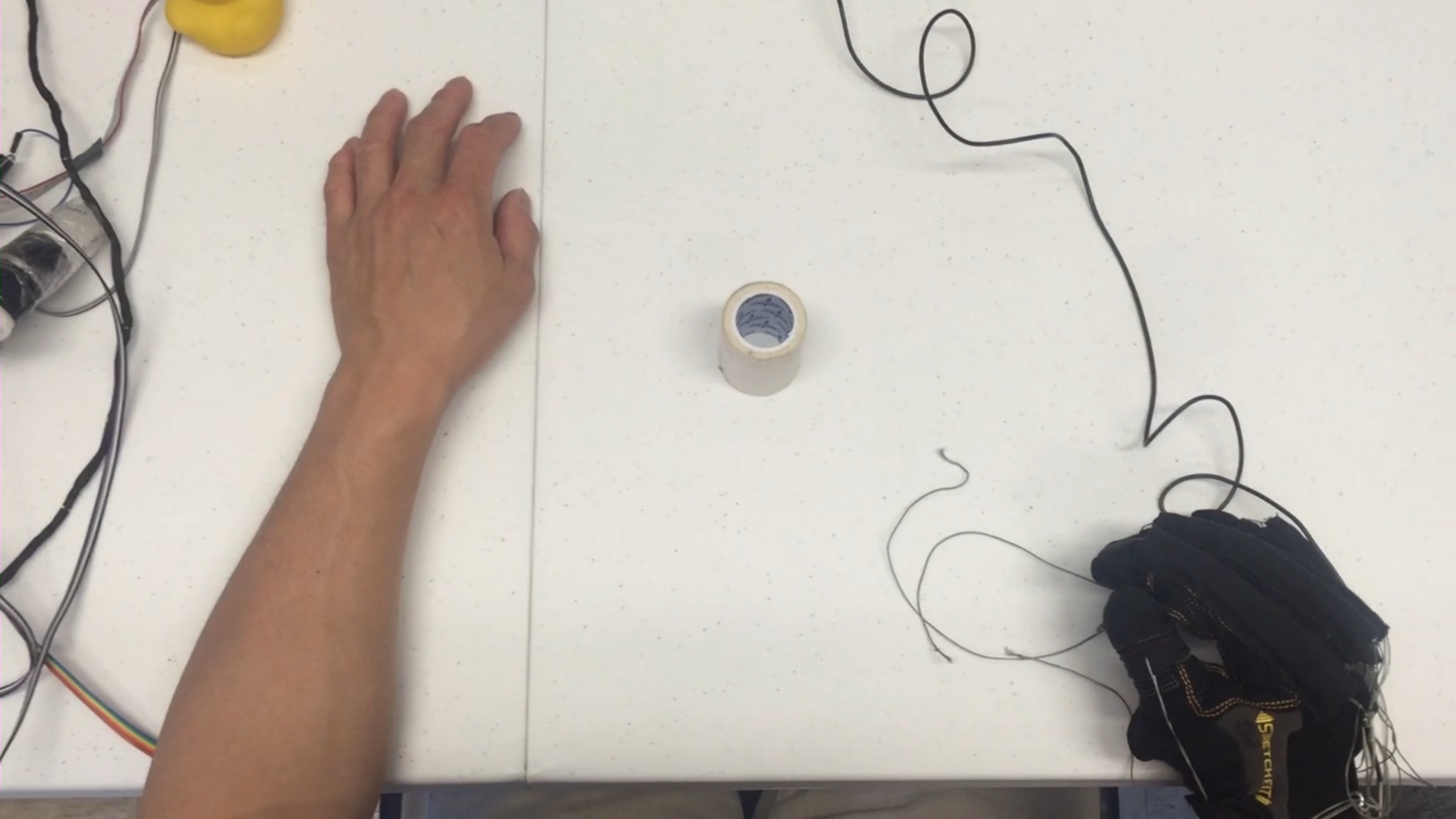}&
\includegraphics[trim=15mm 0mm 0mm 15mm,clip=true,width=0.18\linewidth]{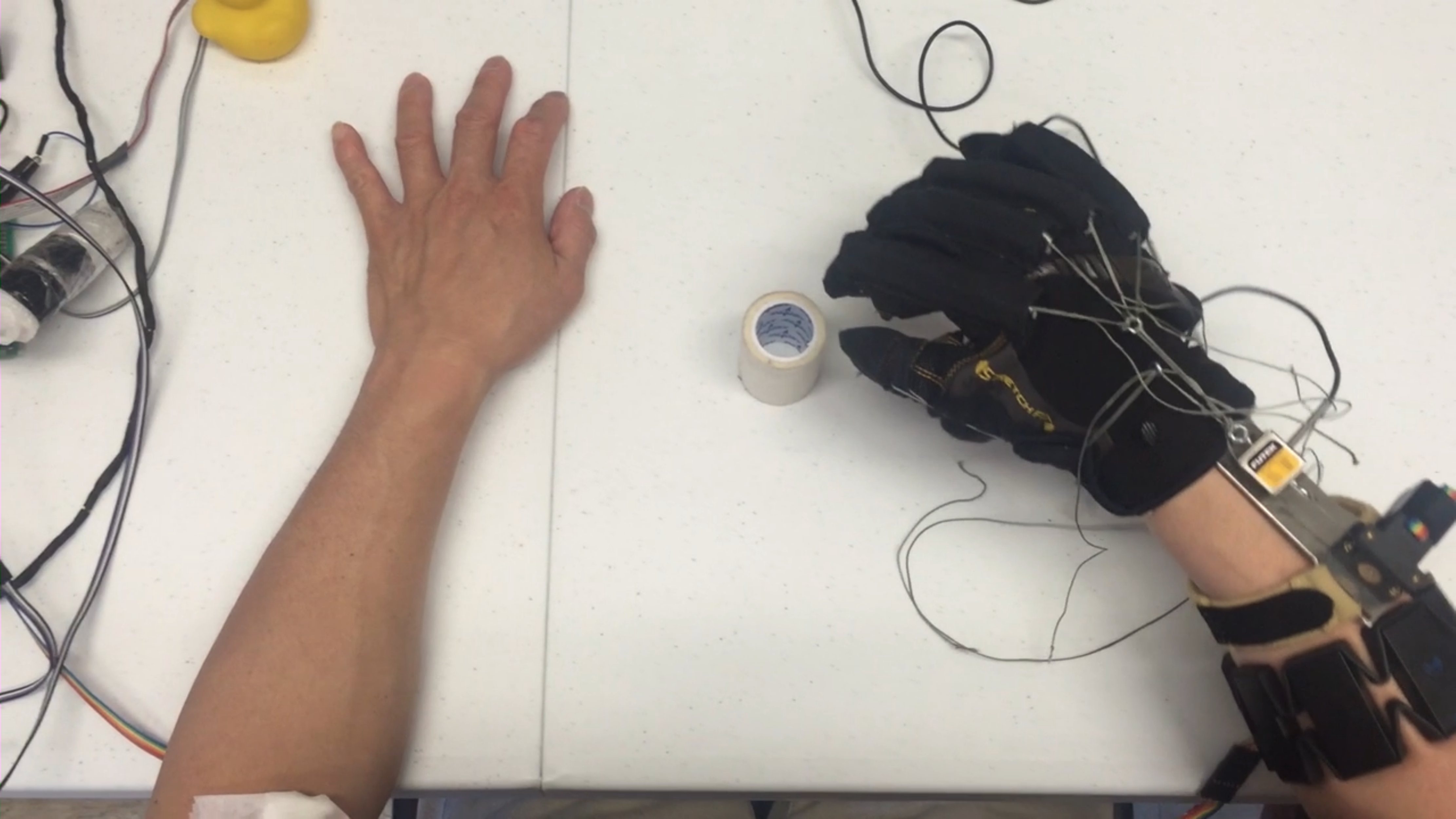} &
\includegraphics[trim=15mm 0mm 0mm 15mm,clip=true,width=0.18\linewidth]{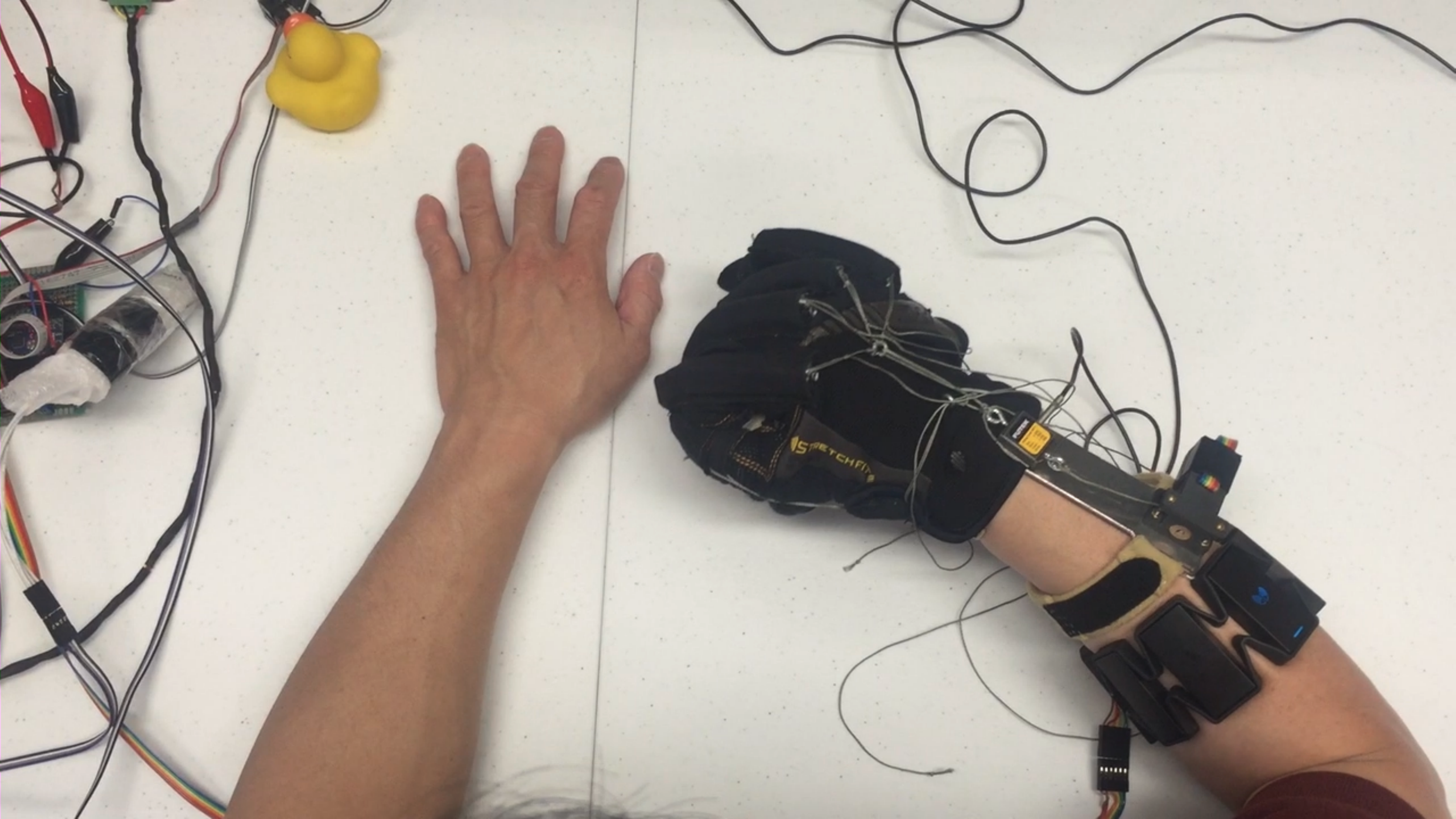}&
\includegraphics[trim=15mm 0mm 0mm 15mm,clip=true,width=0.18\linewidth]{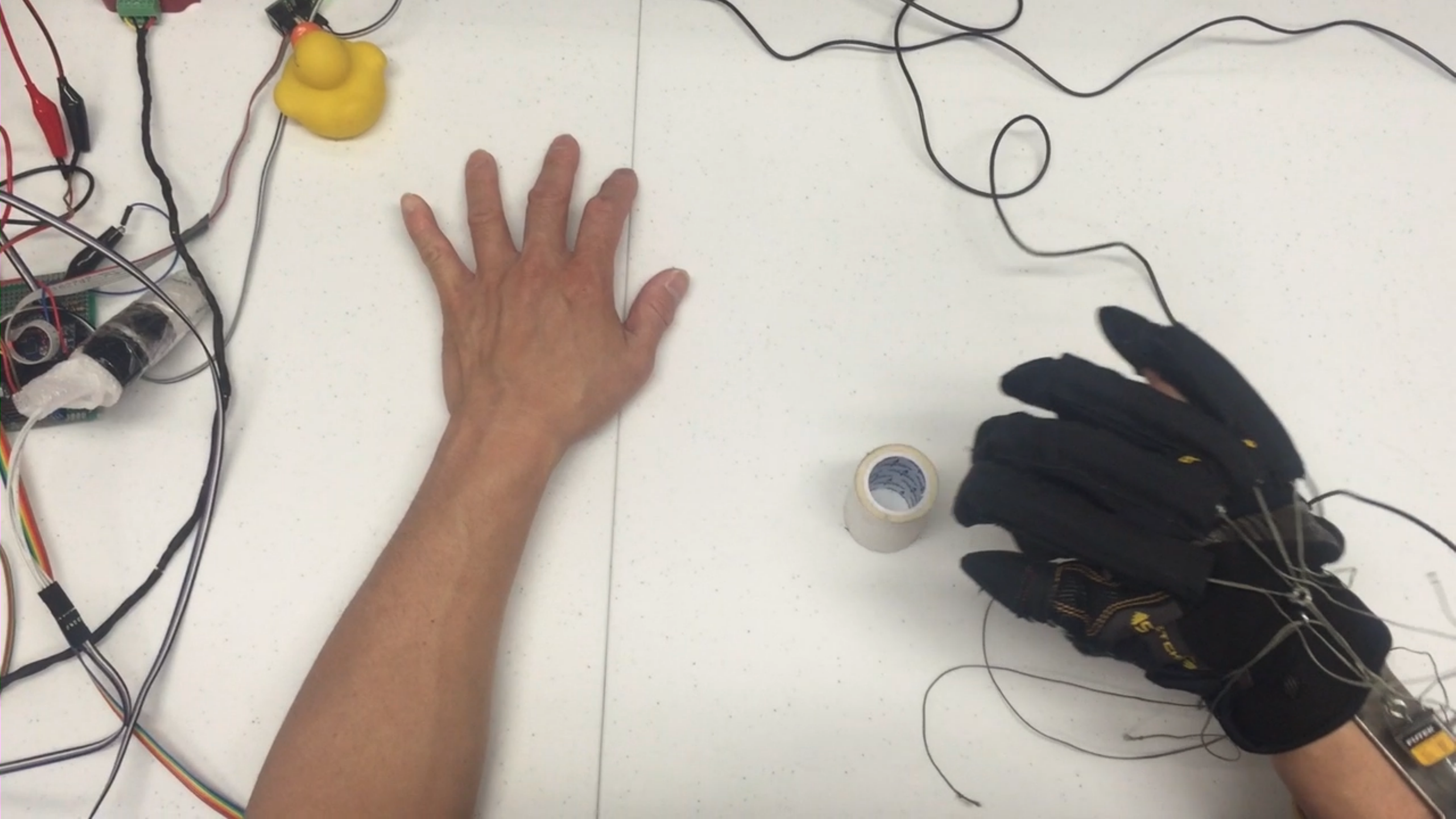} &
\includegraphics[trim=15mm 0mm 15mm 20mm,clip=true,width=0.18\linewidth]{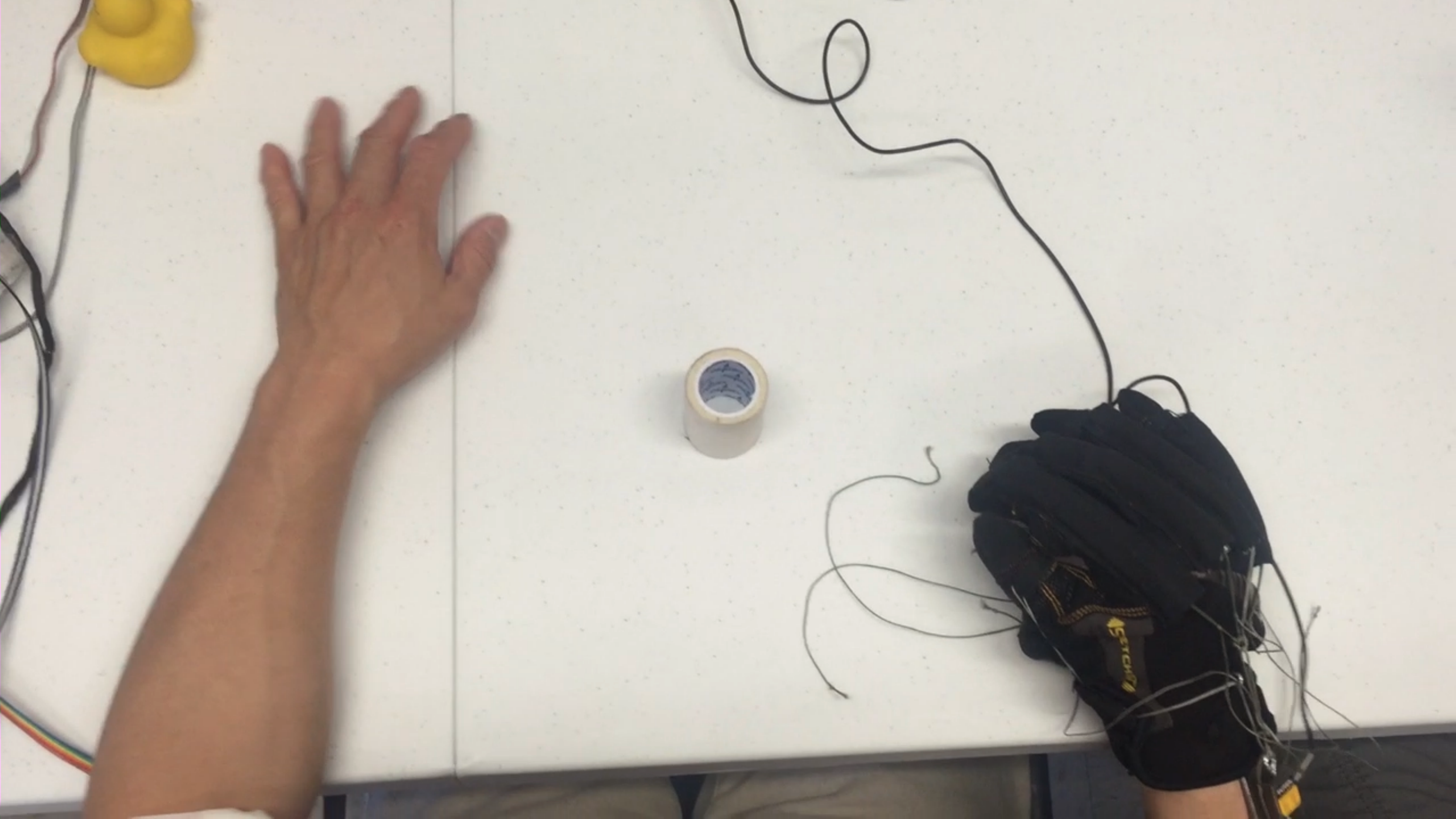} \\[3mm]
\includegraphics[trim=15mm 0mm 0mm 15mm,clip=true,width=0.18\linewidth]{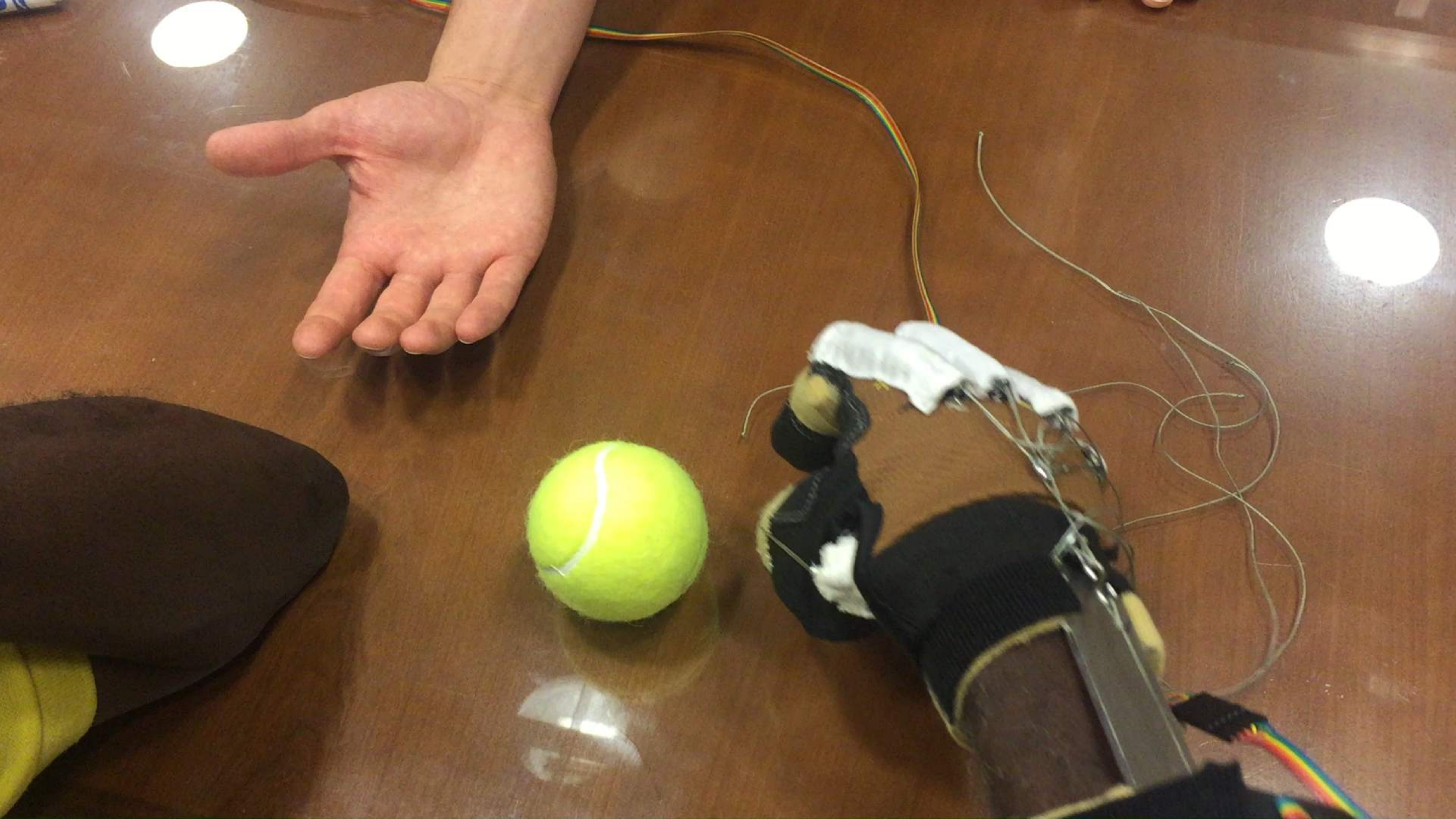}&
\includegraphics[trim=15mm 0mm 0mm 15mm,clip=true,width=0.18\linewidth]{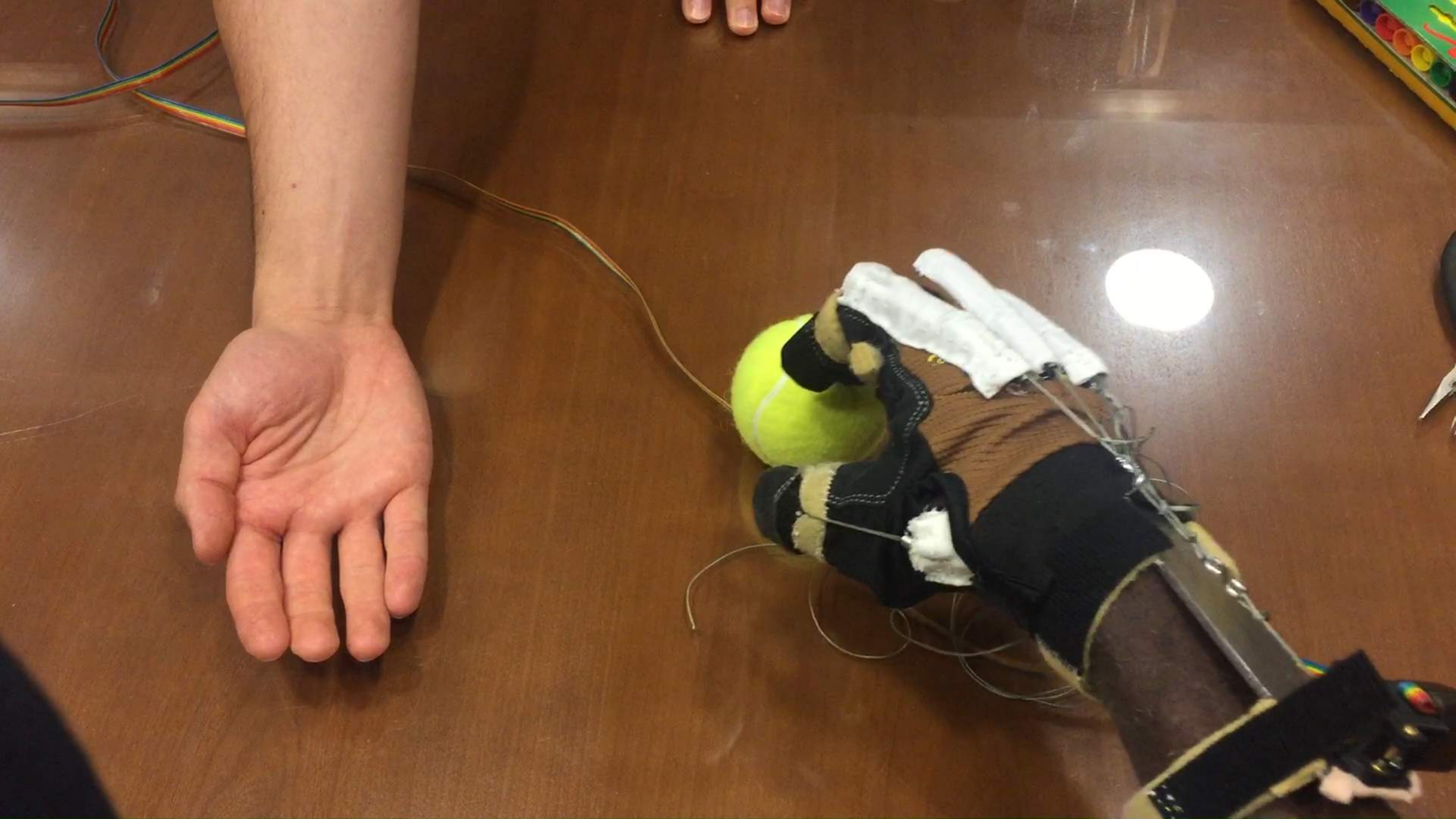} &
\includegraphics[trim=15mm 0mm 0mm 15mm,clip=true,width=0.18\linewidth]{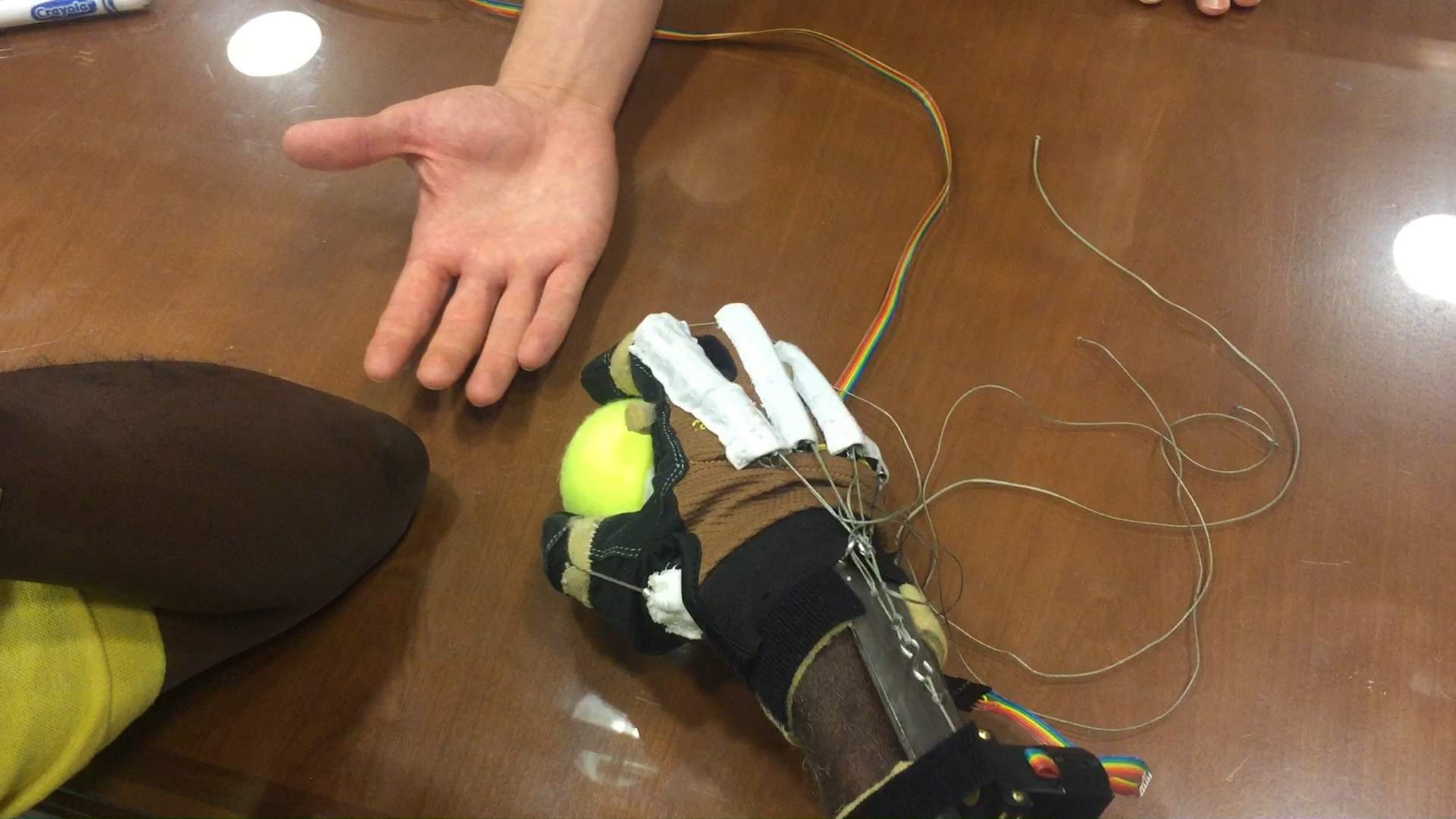}&
\includegraphics[trim=15mm 0mm 0mm 15mm,clip=true,width=0.18\linewidth]{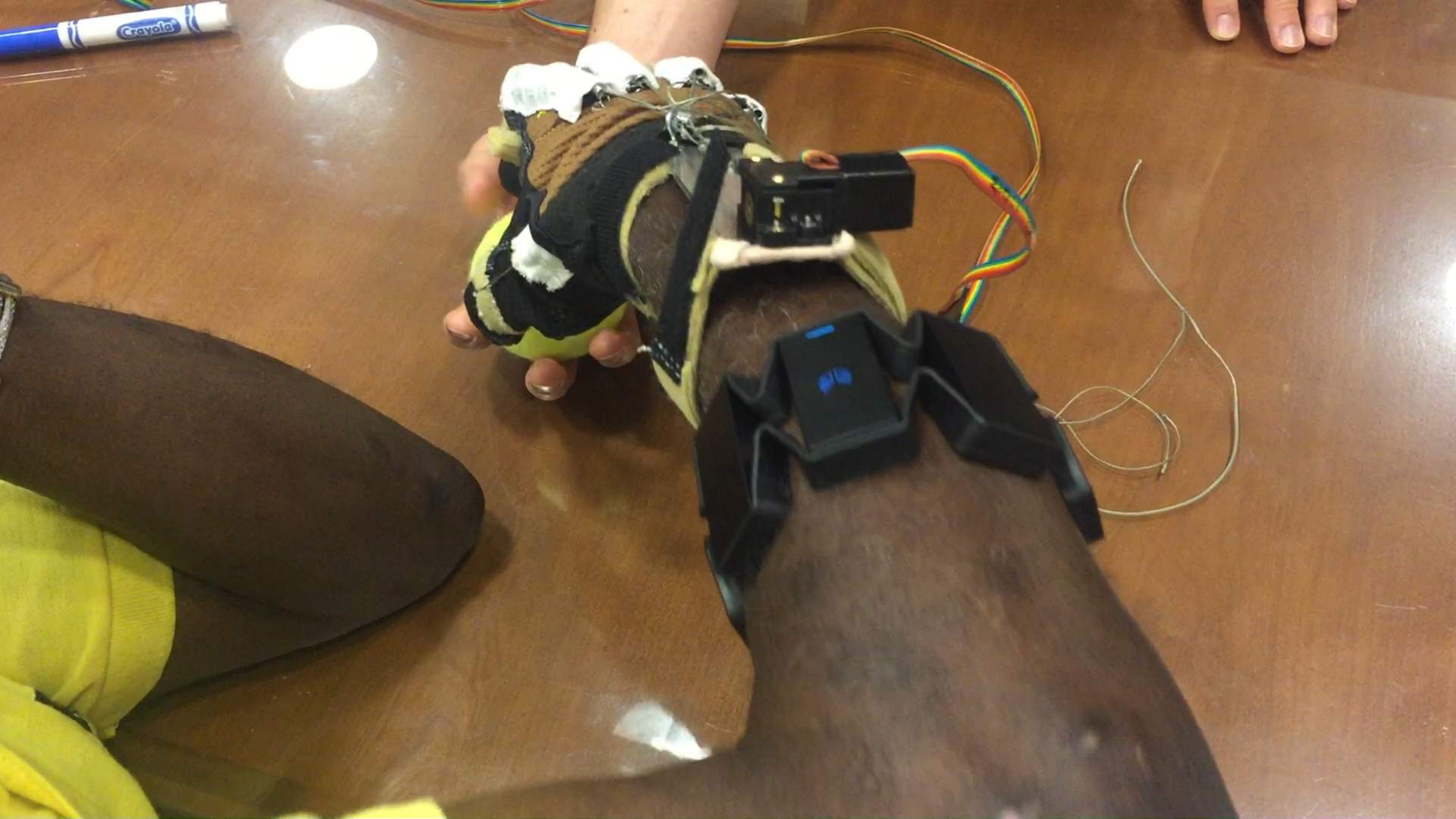} &
\includegraphics[trim=15mm 0mm 15mm 20mm,clip=true,width=0.18\linewidth]{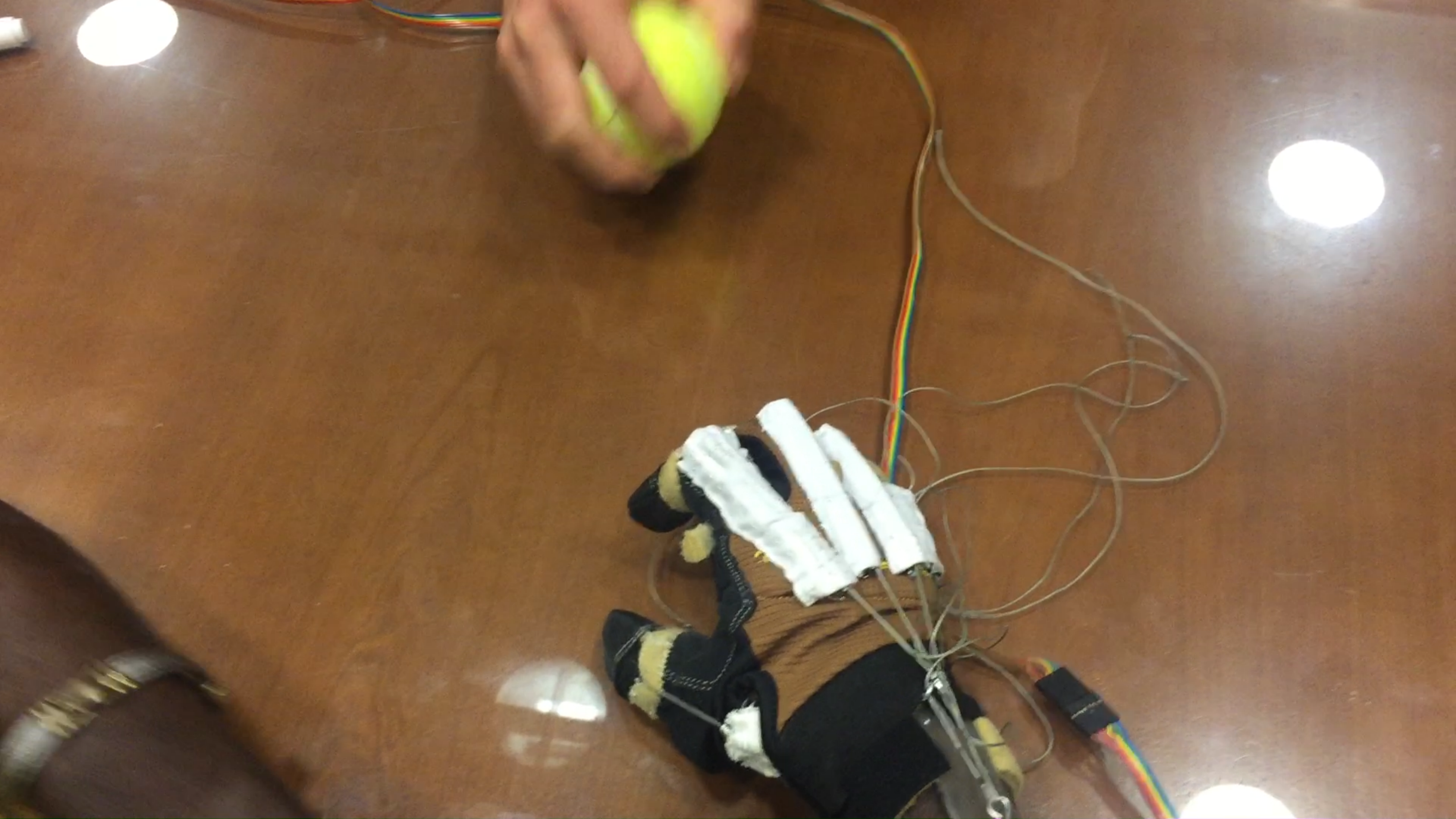} \\
\parbox{30mm}{\footnotesize Subject is at rest. The tendon begins in fully extended state and the hand is closed due to spasticity.} &
\parbox{30mm}{\footnotesize Subject begins grasp. The tendon retracts and the hand opens to enable grasping.} &
\parbox{30mm}{\footnotesize Subject grasps the object. The tendon extends to allow the hand to close over the object.} &
\parbox{30mm}{\footnotesize The subject releases the object. The tendon retracts to open the hand and to allow release of the object.} &
\parbox{30mm}{\footnotesize Grasp is complete and the subject is at rest. The tendon is fully extended.} \\
\end{tabular}
\caption{Illustration of pick and place tasks by 2 subjects. Each row shows a complete task execution by one subject, progressing from left to right.}
\label{fig:pick_and_place}
\vspace{-5mm}
\end{figure*}

\subsection{EMG control during pick and place}

Precise ground truth is difficult to establish when the subject is
performing pick and place tasks because an operator instructing the
user when to begin and end extension would result in unintuitive
grasping. Instead of percent accuracy, we use the number of
correctly executed pick and place tasks as a metric for the pick and
place experiments (both with EMG control and with button control).

We did not do additional training for this set, but used the classifier from the
previous experiment.

During testing, the subject was asked to operate the exotendon device using EMG control 
to pick an object up, move it several inches, and then place it back
down. The details of a complete pick and place motion, as well as the 
exotendon's role in the action are described in Fig.~\ref{fig:pick_and_place}.

Subject B was not included because
sizing issues rendered her unable to grasp
objects. Due to subject fatigue, Subject D was also not
included. Subject A successfully completed 6 of 13 pick and place
attempts. Subject C completed 6 of 6 pick and place movements. We
note that Subject C was higher functioning than the other subjects and
was generally able to complete unassisted hand extension, albeit
with significant difficulty. Nevertheless, the subject reported that
the device provided assistance in hand opening during pick
and place. See Table~\ref{functional_table} for a summary of the
results.

\begin{table}[]
\centering
\caption{Results for Functional Motions (Pick and Place)}
\label{functional_table}
\begin{tabular}{C{1.4cm}|C{1.9cm}|C{1.9cm}}
Subject & Correct Events - EMG Control & Correct Events - Button Control \\ \hline
A       & 6/13        & 3/3            \\ 
C       & 6/6         & 5/5            \\ 
\end{tabular}
\vspace{-6mm}
\end{table}

\subsection{Button control during pick and place}

Before testing, the subject was instructed how to control the
device using their left hand, and allowed to use the control
for several minutes before performing pick and
place tasks. During testing, the subject performed pick and place
on the same object as during the EMG controlled experiment.

Again, Subjects B and D were not included in these results. 
Subject A successfully completed 3 of 3 pick and place
attempts. Subject C successfully completed 5 of 5 pick and place
attempts. See Table~\ref{functional_table} for
a summary of the results.

\subsection{Tendon Forces during Pick and Place Experiments}

\begin{figure}[t]
\setlength{\tabcolsep}{0mm}
\centering
\begin{tabular}{cc}
\includegraphics[trim=15mm 5mm 15mm 6mm,clip=True,width=0.5\linewidth]{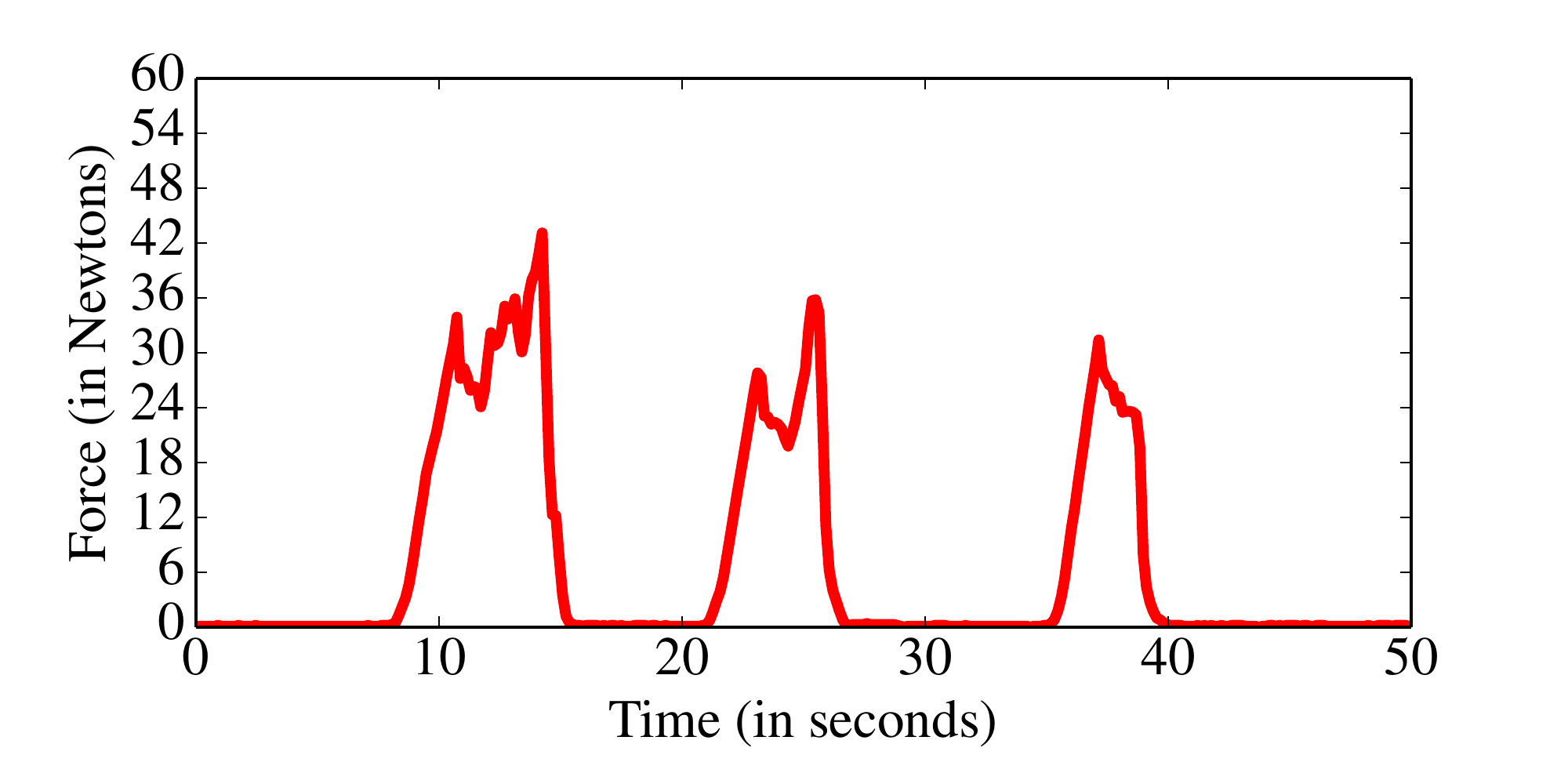}&
\includegraphics[trim=15mm 5mm 15mm 6mm,clip=True,width=0.5\linewidth]{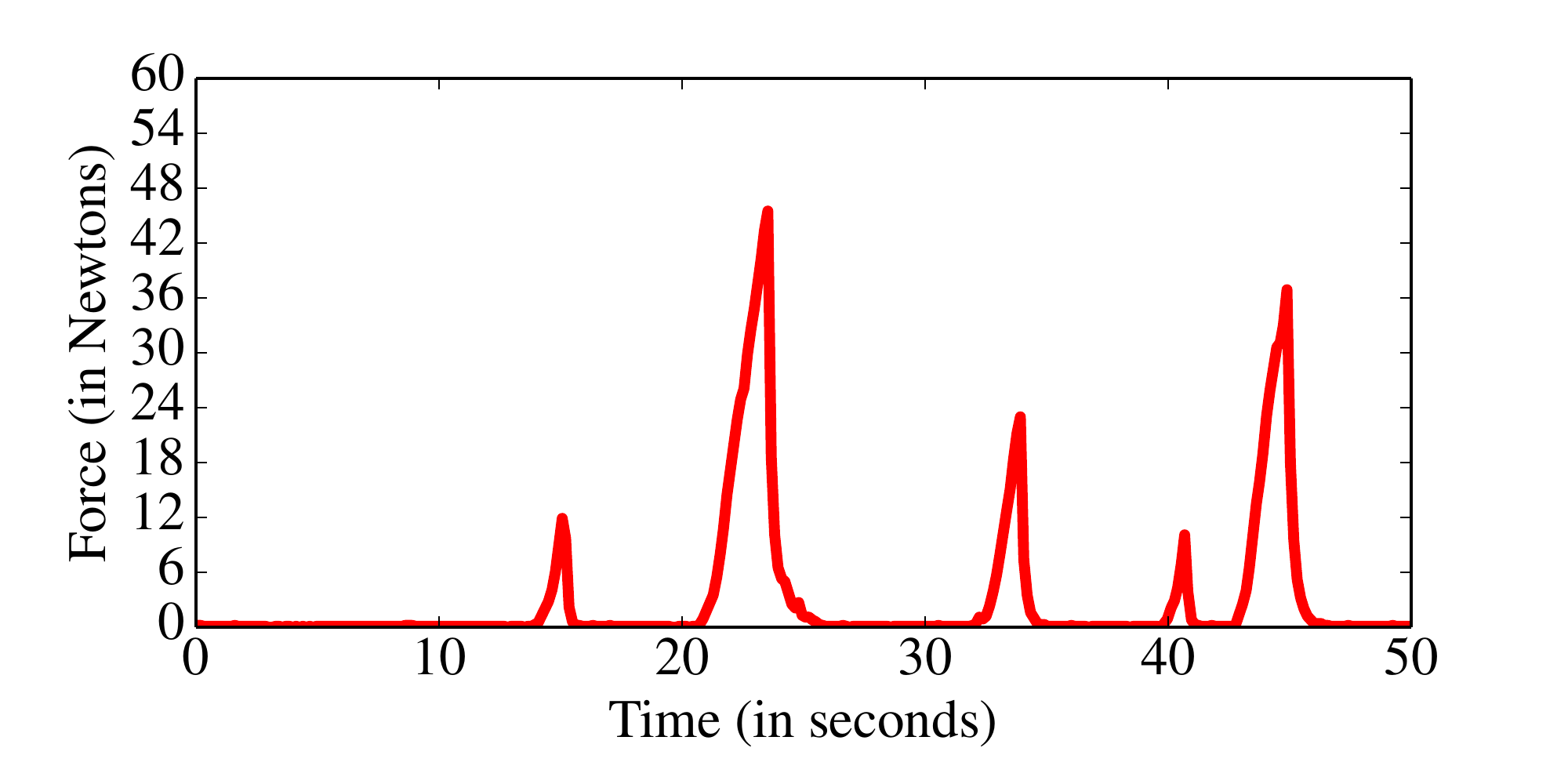} \\
\parbox{40mm}{\footnotesize Force vs. time - button control.} &
\parbox{40mm}{\footnotesize Force vs. time - EMG control.} 
\end{tabular}
\caption{Force results for Subject A - button control (left) and EMG
  control (right). Each peak corresponds to one hand extension
  for grasping.}
\label{fig:force_results}
\vspace{-15pt}
\end{figure}

To assess the ability of the control to correctly interpret
user intent, as well as the level of discomfort caused by operation,
we measured the forces applied on the exotendon network during pick and
place experiments. For the same subject, we compared peak forces
obtained for EMG control versus button control. Our assumption was
that an incorrect interpretation of a ``close'' signal, where the
subject intended to close the hand but the assistive device did not
react appropriately, would result in a spike in the force levels as
the subject would be effectively fighting against the orthosis.

During the button controlled pick and place, the peak force on the
tendon was 53.7N for Subject A and 58.6N for Subject C. During the
EMG controlled pick and place, the peak force on the tendon was 59.6N
for Subject A and 77.6N for Subject C. Fig.~\ref{fig:force_results}
shows an example plot of force vs. time during an EMG controlled and 
a button controlled experiment; each time series contains multiple hand extensions. We measured the forces using a load cell placed in series 
with the exotendon network. The peak 
force for the two controls showed little difference for Subject A, but 
for Subject C, EMG control lead to higher forces. However, there was 
enough variance between force peaks obtained with the same control 
mechanism to suggest that the difference could fall within normal operating
range. Additional testing will be needed to further study this issue.

\section{Discussion}

Overall, our results showed effective pattern classification
performance, to the level of physically enabling functional hand
motion. Still, classification accuracy shows significant room
for improvement. In particular, while the percent accuracy metric was
consistently above 80\% and often above 90\%, the same level of
performance was not achieved in the number of correctly predicted
events. Most of the incorrectly predicted events were the result of
the control not correctly recognizing the change in intention within
the allowed 850ms window, rather than spikes caused by
misclassification in the middle of the event. These delays were
caused in part by the median filter, which uses the past 500ms to
inform the control, thereby adding lag. Another possible cause was
subject spasticity, which made it difficult for the subjects to relax
or close after activating their extensor muscles.

Our results were trained and tested on separate data sets, both of
which were taken from the same patient during the same session. We
would like our trained classifiers to be robust enough to work for the
same patient for different sessions. However, we believe it unlikely
that pattern classification would work between different patients as
EMG patterns between subjects are substantially
different~\cite{lee2011}.

Pick and place experiments controlled by EMG showed lower
accuracy than non-pick and place experiments where the device was
operating. The difference between the 2 types of experiments was that
in the former the subject's arm was engaged in the task, while in the
latter the forearm was simply resting on the table. We hypothesize
that, because of the stroke subjects' abnormal coactivation,
a classifier which was trained while the forearm was resting
on the table is confused by elbow extension during a 
grasping motion. We hope to compensate for this effect in future
iterations by altering our training protocol to include training data
both when the arm is resting and when the arm is extended.

As our work is eventually intended as a take-home device, the level of
automation of the training is an important consideration. In this
study, an operator was required to provide the
training set with ground truth while instructing the patient to try to
open or close. The operator also used the button control to implement
the training protocol when required. In the future, the above
responsibilities could be transferred to a user
interface using visual cues instead of verbal commands, instructing
the patient when to try to open and close, and programmed motor
actions.

\section{Conclusions and Future Work}

In this paper, we have shown that a EMG based pattern classification
control of an exotendon device can enable functional movement in a
stroke survivor. Our control achieves high accuracy during
non-functional open and close hand motions, and can enable functional
motions, like pick and place. The pattern classification technique
allows the use of commodity devices which are easy to don, as there is
no need to place sensors on specific muscles. Our control is intuitive
and does not require an extended period of training. Our study shows
that functional movement can be enabled by EMG control in wearable
devices.

In the future, we would like to:
\begin{itemize}
\item Make our control robust to donning and doffing without having to take training sets each session.
\item Make our control more robust to the abnormal coactivation
  initiated in stroke patients during functional tasks which require
  elbow extension.
\item Include more hand movement patterns into the EMG 
  classifier. We would like to differentiate between the 2
  whole hand movement patterns of hand extension and of
  metacarpophalangeal (MCP) flexion / interphalangeal (IP)
  extension~\cite{park2016}. This would provide the user with
  functional motion assist for multiple types of grasps.
\item Explore the idea of training users to
  produce movements which are more easily distinguishable for a
  classifier~\cite{powell2013}. Although presented in the field of
  prosthetics, it would be useful for orthotics because abnormal
  coactivation in stroke patients makes classification difficult.
\end{itemize}
We hope that our control, and future iterations, will inform the
development of a wearable orthosis that can be used outside of
clinical settings.

\bibliographystyle{IEEEtran}
\bibliography{bib/prosthetics,bib/other_control,bib/orthoses,bib/stroke,bib/ML}

\end{document}